\documentclass{article}

\usepackage{arxiv}

\usepackage[utf8]{inputenc} % allow utf-8 input
\usepackage[T1]{fontenc}    % use 8-bit T1 fonts
\usepackage{hyperref}       % hyperlinks
\usepackage{url}            % simple URL typesetting
\usepackage{booktabs}       % professional-quality tables
\usepackage{amsfonts}       % blackboard math symbols
\usepackage{nicefrac}       % compact symbols for 1/2, etc.
\usepackage{microtype}      % microtypography
\usepackage{lipsum}		% Can be removed after putting your text content
\usepackage{graphicx}
\usepackage{amsmath}
\usepackage{array}

\title{User-Oriented Smart General AI System under Causal Inference}

\date{} 					% Or removing it

\author{ \href{https://orcid.org/0000-0001-7431-1619}{\includegraphics[scale=0.06]{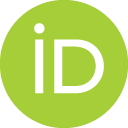}\hspace{1mm} Huimin Peng}\thanks{
		Thank you for all helpful comments! Feel free to leave a message about comments on this manuscript. In case I did not receive email, my personal email is \texttt{974630998@qq.com}. Thanks to github.com/kourgeorge/arxiv-style for this pdf latex template.} \\
	\texttt{peng.huimin.pennie@gmail.com} \\
}

\begin{document}
\maketitle

\begin{abstract}
General AI system solves a wide range of tasks with high performance in an automated fashion. 
The best general AI algorithm designed by one individual is different from that devised by another. 
The best performance records achieved by different users are also different.
An inevitable component of general AI is tacit knowledge that depends upon user-specific comprehension of task information and individual model design preferences that are related to user technical experiences. 
Tacit knowledge affects model performance but cannot be automatically optimized in general AI algorithms. 
In this paper, we propose User-Oriented Smart General AI System under Causal Inference, abbreviated as UOGASuCI, where UOGAS represents User-Oriented General AI System and uCI means under the framework of causal inference.
User characteristics that have a significant influence upon tacit knowledge can be extracted from observed model training experiences of many users in external memory modules.
Under the framework of causal inference, we manage to identify the optimal value of user characteristics that are connected with the best model performance designed by users. 
We make suggestions to users about how different user characteristics can improve the best model performance achieved by users. 
By recommending updating user characteristics associated with individualized tacit knowledge comprehension and technical preferences,
UOGAS helps users design models with better performance. 
\end{abstract}

% keywords can be removed

%\section{Introduction}
%\label{introduction}

\section{General AI}
\label{General-AI}

%purpose of general AI
General AI has wide applications in robotics, image processing, natural language processing, etc. 
General AI aims to design capable algorithms that provide automated solutions to diverse tasks. 
%why develop general AI
Developments in general AI are significant contributions to improvement upon current deep learning models. 
With large-sample and high-dimensional data to process, training deep models from scratch is time-consuming and requires user attention into model specifications for each diverse task. 
%acceleration of current deep models, improve efficiency, enable applications
General AI methods guide model training with meta-learner for higher efficiency while widening applicability to more tasks. 
In current deep models, introducing general AI components contributes to better training efficiency and generalization capability.
%applied to wide range of tasks
%user-friendly, one model for all tasks, make general AI readily available to everyone
Since general AI is designed to be automated, it solves similar tasks efficiently with little additional training on each new task. 
General AI alleviates the constant labor of doing repeated tasks and assists humans on unseen tasks by recommending useful suggestions from previous experiences.  
%automatic data processing, save trouble
With general AI, users only need to learn how to use one system to solve a wide range of tasks. 

General AI is a promising domain with developments from multiple perspectives each offering unique contributions to the construction of concrete algorithms.
%current developments in general AI
AlphaGo Zero \cite{Silver2018} outperforms the best human players in chess, shogi and Go games. AlphaGo Zero takes explicit game rules as input and consumes only hours of training time. 
%alphazero, ai-ga, autoformalization 
AI-GA (AI-Generating Algorithm) \cite{Clune2019} finds feasible solutions to tasks that contain varying complex environments and cannot be solved well by training deep models from scratch. 
Autoformalization \cite{Szegedy2020a} conducts causal reasoning using automated deep models and is applicable to diverse reasoning tasks on natural language processing. 
For an unseen task, general AI methods perform self-improvement efficiently based upon former training experiences.
In conclusion, a general AI system contains two components: a general architecture that is suitable for diverse tasks, and a self-improvement mechanism that updates deep models efficiently to solve new tasks.

%risks
General AI does not contain risk with respect to uncontrolled model complexity growth or evolution into strong AI. 
%uncontrolled complexity growth, not possible in nature, hyper-parameters vary in cycles
First, hyper-parameters in deep models are designed to vary in bounded cycles which constrains model complexity growth \cite{Peng2021}.
%not possible in algorithms as well, algorithms do not exhibit such growth potential, all algorithms generally fail when complexity is very high, algorithms are not that scalable
Second, algorithms are in essence not scalable to task complexity growth. 
There does not exist algorithm or architecture that is scalable to overwhelmingly fast and unbounded task complexity growth.
For example, statistical models developed for low-dimensional data do not scale well to high-dimensional task data.
In meta-RL, as task environment becomes more complicated, trained models fail to provide sensible solutions.  
Even with such a long history of evolution, humans cannot solve all problems on earth. 
%Algorithms generally fail when task complexity increases above thresholds.
Nice properties of models are subject to assumptions on task data and parameters, which will be violated as model complexity grows in an unbounded manner.   
%general AI solves wide range of tasks, but does not scale well with respect to task complexity growth
Although general AI models are claimed to be as general as possible, they are subject to assumptions. 
For example, neural network models restrict the input image data to be of the same dimension, i.e. same number of pixels, same data shape. 
Images in the same dataset are of the same dimension so that the same feature extraction module is applicable to all. 
Since models are restricted by assumptions, general AI does not possess the risk of developing into strong AI. 
Pursuing general AI brings us closer to more general deep models that are less likely to over-fit and improves model training efficiency that enables wider applications of deep models.  

\section{User-Oriented Design}
\label{User-Oriented-Design}

%human affects tools, designs tools, update tools
Humans design tools to bring convenience and to improve efficiency, and then update tools to solve new challenges. 
%tools also affects human, allocate more attention to disadvantage of tools, more free time or less, relying upon tools changes human behavior, focus attention upon problems that tools cannot handle
With the help of tools, humans allocate more attention to drawbacks of tools and problems that current tools cannot handle. 
%human-algorithm interaction, human design algorithms, algorithms change human behavior, similar
Humans interact with general AI systems where designed algorithms affect human behaviors and vice versa. 
%why user-orientation
To promote the application of general AI systems, algorithms are designed in a user-oriented way where users take recommendations from algorithms to design models with better performance. 

%inevitable tacit knowledge required to process tasks
In the process of solving diverse tasks, humans comprehend task information to acquire tacit knowledge based upon which algorithms are designed.  
%tacit knowledge is user-based from human comprehension
Tacit knowledge is user-specific and hinges upon model training experiences and technical abilities of users. 
%
%algorithms have two parts
General AI algorithms include two components.
\cite{Peng2021}
%human-interpretation, tacit knowledge based, ad-hoc, hyper-parameters
The first part is based upon tacit knowledge from user-specific interpretation of task information and individual technical preferences.
It is thought that tacit knowledge cannot be explicitly optimized in general AI algorithms.
%automatic parts determined through optimization process, determined within algorithms
The second part includes all unknown parameters that are automatically determined through optimization within algorithms. 
%
%
%for example,
%the process of choosing varying hyper-parameters in AutoML, neural network model, we may choose.... these are decided by human from user experiences
For example, in AutoML, the set of hyper-parameters to be optimized and the way in which neural network models evolve are devised by users. 
The set of variable hyper-parameters may include width and depth, activation functions, the number of nodes in output layer, shortcut paths, etc.
Neural network contains many variable elements that can be covered in the optimization of hyper-parameters. 
It is up to users to decide which variable elements are taken into consideration in search for the optimal network structure. 

%objective functions are ad-hoc, in automatic navigation, objective can be minimization of travel time, minimization of consumed gasoline, minimization of distance travelled, subject to constraint such as mid-destination, avoid highway, avoid toll roads
Objective functions and constraints in optimization within algorithms are ad-hoc and determined by tacit knowledge from user-specific interpretation of task information.
In automated car navigation, users may specify objectives as minimization of time spent, minimization of distance travelled, minimization of gasoline consumed, etc.  
Users may impose constraints such as avoiding highway, avoiding toll roads, mid-destinations that the car must stop by, etc. 
%
%scope of applied models, limited by human comprehension of machine learning, possess knowledge to use models, understand a range of models but not all of them, knowledge scope, education
Users tend to apply models that they are more familiar with.
Individuals acquire different histories of model training experiences and knowledge scopes in machine learning. 
Choices of specified models are also user-specific and rely upon such tacit knowledge.

%
%the goal of general AI is to develop a general algorithm good for lots of tasks so that non-experts can use general AI to solve all tasks, more accessible to human
The primary goal of general AI is to develop a general algorithm applicable to a wide range of tasks so that experts or non-experts can benefit from general AI and learn only one system to solve all tasks. 
%user-orientation makes it certain that general AI suits user better, by performing optimization conditional upon user characteristics, provide suggestions to users that can outperform the best AI algorithm that current user knowledge can possibly design
User-oriented design of general AI guarantees that devised systems better suit user-specific requirements.
User-oriented systems perform optimization conditional upon user characteristics and
provide individualized suggestions to improve the best AI algorithm that current users can devise. 
%provide suggestions to the kind of tacit knowledge human need to better design AI algorithm in this task
Modifications recommended by user-oriented general AI systems update user characteristics which have an influence upon user-specific tacit knowledge infused within designed AI algorithms. 

%three functions 
In external memory modules, model training experiences of many users and many tasks are collected and ready to use. 
User-oriented algorithm design 
consists of three steps \cite{Unknowng}.
%decide which user characteristics are causal effects upon the best general AI performance achieved by that user individual
First, dimension reduction is applied to identify the set of user characteristics that affect tacit knowledge and are truly relevant to the performance of designed algorithms. 
Causal inference is provided upon user characteristics to analyze effects upon the best performance achieved by that individual. 
%how varying these user characteristics will contribute to performance improvement
Second, it is crucial to determine how varying these user characteristics will contribute to improvement in performance. 
%make these suggestions on user characteristics to the user so that user can make adjustments to make individualized better design
Third, user-oriented systems issue recommendations upon the optimal values of user characteristics to achieve the best performance.
Reliability and robustness of these suggestions are analyzed and users may choose to accept proposed modifications or not. 
Users may make these adjustments for self-improvement and to reach better individualized design.

%general AI solves all tasks with highest performance, ideally
Ideally, general AI seeks to tackle diverse tasks with the highest achievable performance \cite{Peng2021}. 
%measure is to maximize the proportion of tasks general AI can solve and maximize the performance of AI on these tasks, maximize on both sides
An appropriate performance-based objective of a general AI algorithm is the joint maximization of (1) the proportion of tasks an algorithm can solve and (2) the average performance of an algorithm on these tasks. 
%different user experiences are different
%in reality, same model shows different performances on different tasks, most suitable for a particular task
In reality, there does not exist such general AI algorithms that can solve all tasks with perfect performance. 
In some cases, algorithms show the best performance on one dataset but not the best performance on others. 
In some scenarios, algorithms solve a wide range of tasks with satisfying performance but do not show the best performance on these tasks than other algorithms.

%general AI is designed by human user
General AI methods are designed by humans and infused with user-specific tacit knowledge concerning tasks. 
%affected by human user
%offer suggestions to user so that user re-design general AI algorithm to be better than the best AI that user can design
User-oriented general AI systems offer suggestions for users to update their own characteristics and to re-design general AI algorithms to be better than the former best algorithms devised. 
%know weaknesses of user, choice bias of user that adversely impact AI performance
User-oriented general AI systems accumulate past model training experiences of many users with different characteristics from diverse background. 
By calibrating the current user against others through channels of meta-learning and causal inference, user-oriented general AI systems identify the set of user characteristics affecting performance of designed models. 
After locating weaknesses of current users, 
%provide recommendation of better model specifications to user
user-oriented general AI systems provide recommendations pertaining to the optimal values of user characteristics that are beneficial to the performance of designed models.

Under this paradigm of human-AI interaction, users modify values of user characteristics to be optimum, integrate user-specific tacit knowledge into algorithms, and optimize algorithms for the best model performance.
Traditionally user-specific tacit knowledge is not considered in automated optimization within algorithms. 
However, in user-oriented general AI systems, user-specific tacit knowledge is optimized by calibrating the current user against others in terms of user characteristics and model training experiences through meta-learning and causal inference. 
Under the framework of causal inference, user-specific characteristics are optimized and issued as suggestions to users in the hope that modified user-specific tacit knowledge will lead to better model performance.
In the context of causal inference, there is no global optimum that works for everyone. 
Every individual has his own personalized optimum that works best only for himself. 
For example, recommendations issued to students from statistics background may be to take more computer science courses. Recommendations issued to students from computer science background may be to take more programming courses.
These recommendations are user-specific and hinge upon other individual features. 
The objective of these recommendations is to improve the best performance of models designed by the user. 

\section{Causal Inference}
\label{Causal-Inference}

%association and causality
Association and causality are different, where causality is often regarded as a more fundamental concept than association. 
%holding all else fixed, merely change one factor, brings the change, so that this factor is the cause, the change is the result, causality established
To describe causality, holding all else fixed, we change only one factor and observe consequences, so that this factor is the cause variable, observations are the effect variables. 
%one factor increases and the other one increases as well, from long observation, these two factors tend to vary in the same direction, they are associated
To depict association, from many observations, factors tend to vary in the same direction so that they are positively associated.
On the contrary, from many observations, factors tend to vary in the opposite direction so that they are negatively associated.
Ideally, to demonstrate causal relation and to measure magnitude of causal effect, we need to create scenarios where varying only cause variables leads to changes in effect variables. 

%ideally, causality is more essential than association. links in neural network are association, but can be causality as well by design
In most algorithms, association is analyzed between variables, not causal relation. 
Causality can be represented with mathematical reasoning and knowledge hierarchy. 
On the other hand, causal inference provides statistical inference to answer whether significant causal relations exist between variables. 
To conduct causal inference, we need observed data including cause variables, effect variables and all confounding variables. 
Under restricted assumptions, we can use these observational data to measure the impact of causal relations upon outcomes.  
These strong assumptions can be validated with graph network models which estimate the dependence structure between variables. 

%causal inference assumptions
Causal inference is subject to many assumptions concerning confounding variables, effect magnitude of causal relations, heterogeneity of causal effects on individuals, etc \cite{Unknown}. 
%inference on causality is based upon these model assumptions
%observational studies
In practice, values of effect variables hinge upon cause variables, other confounding variables, etc. 
For a set of cause variable values, effect variables take a corresponding set of values. 
Conditional upon a set of cause variable values and other confounding variables, effect variable values follow a corresponding probability distribution.
To measure the effect of causal relations, it is assumed that there is no other unobserved confounding variable influencing effect variable values. 

%optimization after establishing causality
%\cite{Dawid2000}
%counterfactual
%we only observe what happen due to past causalities
Ideally, on the same individual, we vary cause variable values and observe the corresponding effect variable values to obtain a full diagram of causal effect. 
In some cases, observed causal effects are non-invertible and we cannot observe effect variables at other values of cause variables. 
For each individual, we only observe the effect variable values given one set of cause variable values. 
After a set of cause and effect variable values is observed, the individual is influenced and is not in desirable conditions to take another set of cause variable values. 
Counterfactuals and potential outcomes are the potential effect variable values that would be observed on the same individual given other sets of cause variable values \cite{Dawid2000}. 
Counterfactuals and potential outcomes are not observed and only exist in theory to construct unbiased estimators of causal effect upon individuals. 

%we cannot observe outcomes of all options 
%counterfactuals are outcomes of the other options that do not actually happen so that we do not observe their outcomes, potentially their outcomes exist and are regarded as counterfactuals
%we can choose the best option to implement only after we observe outcomes of all options, we choose the option with the best outcome
%we see the future and can make the best choice for now
It is not possible to observe the future outcomes, change past cause variable values, re-observe the future outcomes on the same individual, and finally find the optimal cause variable values corresponding to the best future outcomes. 
%but we dont get to observe future
%
%we estimate outcomes of all options for each individual
Instead we observe cause variable values and the corresponding effect variable values for a large number of individuals. 
%we observe lots of individuals, they make different choices, select different options, observe outcomes for all of them, their characteristics
After considering all confounding variables, we analyze outcomes of different cause variable values and estimate the magnitude of causal effect. 
For each individual, the optimal cause variable value is different and depends upon confounding variables. 

%\cite{Unknowng}
%precision medicine
An important application of causal inference is in precision medicine to identify the individualized optimal treatment decision \cite{Unknowng}. Cause variables are treatment options and effect variables are treatment outcomes. 
%causal inference framework
%provide individualized optimal treatment regime
The optimal treatment decision for one patient is different from that for another patient especially when patients are vastly different. 
Patient characteristics are confounding variables and may affect both cause and effect variables.
Under causal inference framework, it is assumed that all such confounding variables are observed and considered in estimating the magnitude of causal effect on each individual. 
%prescribe medicine that is optimized for each individual
With causal inference in place, doctors prescribe the optimal treatment conditional upon all relevant patient characteristics. 
%best choice for you may not be best choice for others, because you are different from others

In user-oriented design of algorithms, cause variables are the user characteristics that causally affect the best model performance achieved by the user. 
Effect variable is the best performance of models devised by the user.
Other individual features that affect both cause and effect variables are observed and considered when estimating the optimal value of cause variables for each user.
Based upon specific conditions of users and tasks, the optimal user characteristics corresponding to the best performance of designed models are identified. 
Such user characteristics determine tacit knowledge from human interpretation of task information and scopes of user technical experiences, which affect the performance of devised models. 
%options-outcomes-individual characteristics
%

\section{Optimization of User-Oriented Design}
\label{Optimization-of-User-Oriented-Design}

In the large dataset from external memory modules, we observe model training experiences of many users for a wide range of tasks.
User characteristics and model performance are considered to find the optimal value of user characteristics that leads to the best performance of models designed by the user. 
Given individual features covering user characteristics and all confounding variables, causal relations between user characteristics and the best performance of user-designed models are studied. 
Recommendations concerning the optimal user characteristics corresponding to the best model performance are proposed to users from user-oriented smart general AI systems. 

\begin{figure}[htpb]
	\centering
	\includegraphics[width=0.85\linewidth]{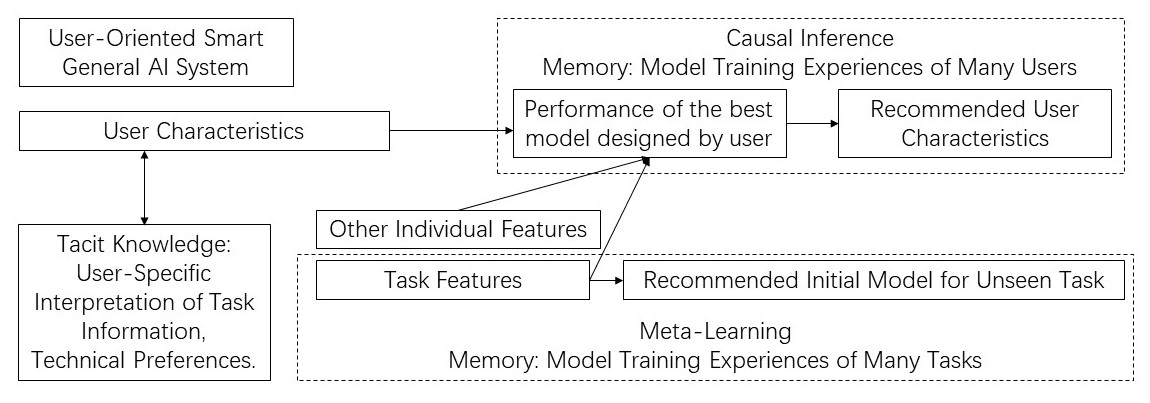}
	\caption{UOGAS: User-Oriented Smart General AI System.}
	\label{p1}
\end{figure}

Figure \ref{p1} presents a sketch on UOGAS (\underline{u}ser-\underline{o}riented smart \underline{g}eneral \underline{A}I \underline{s}ystems). 
UOGAS consists of two main components: a meta-learning channel and a causal inference channel. 
First, the meta-learning structure uses model training experiences of many tasks to propose good initial models for an unseen task. 
There are many algorithms in meta-learning that conduct efficient model self-improvement to reach satisfying performance on new tasks.
Second, the causal inference architecture uses model training experiences of many users to identify the optimal values of user characteristics that lead to the best performance of models designed by this user. 
These user characteristics determine the way in which users interpret task information to extract tacit knowledge exploited to formulate the general AI algorithms for these tasks. 
Furthermore, these user characteristics are user-specific cause variables of model performance and manipulating values of these user characteristics should improve model performance.
These user characteristics are user-specific and cannot be accounted for elsewhere in algorithms. 
In addition, more components can be integrated into UOGAS to fully exploit data accumulated in memory modules. 

User characteristics represent tacit knowledge which is user-specific interpretation of task information and user preferences for applying more familiar models. 
As a result, user characteristics affect performance of the best model designed by the user. 
Under the framework of causal inference, assume that all confounding variables (between user characteristics and the best model performance) are observed in other individual features and task features.
User characteristics, other individual features, task features and performance of the best model designed by the user are calibrated against items in the external memory bank, where model training experiences of many users are saved. 
Conditional upon other individual features and task features, the optimal value of user characteristics corresponding to the best performance of models designed by the user is provided as recommendations. 
To be more user-friendly, it is up to the users to decide whether to accept the updates suggested by the recommendations or not. 
On the other hand, under the framework of meta-learning, task features are compared with items in memory bank, which records model training experiences of many tasks \cite{Peng2021,Peng2021a}. 
An initial model is recommended to the user for an unseen task.   

\subsection{User-Oriented Hyper-Parameters}
\label{User-Oriented-Hyper-Parameters}

%for example, in autoML
To begin with, autoML is used as an example to illustrate hyper-parameters that represent user-specific components such as tacit knowledge obtained from task information and technical preferences of users. 
Since users tend to apply models that they are more familiar with, algorithm choices are bounded within individual technical skill sets. 
Making sense of task information relies upon past model training experiences of users, current requirements, situations and environments, etc. 
These components are directly determined by users and cannot be automatically optimized in algorithms. 

%three parts of parameters
%model y=f(x,theta)
AutoML searches over the space of all variable hyper-parameters to find the optimal neural network structure with the best performance on this task. AutoML is a general AI architecture since it is applicable to all tasks that can be solved with neural network models. 
Denote an autoML model as 
\[
y=f(x,\theta)+\epsilon,
\]
where $y$ is the data annotation, $f$ is the neural network model, $x$ is the input data, $\theta$ is the unknown parameter of interest, and $\epsilon$ is the random error. 
$\theta$ comprises three components: 
\[
\theta=(\theta_1,\theta_2,\theta_3).
\]
%theta1
%human-interpretation ad-hoc, determines model architecture, objectives, varying components in models, hyper-parameters to optimize
$\theta_1$ represents the hyper-hyper-parameters that set variable elements in the model and the rules in which neural network structures evolve.  
$\theta_1$ determines the set of hyper-parameters to be optimized in the neural network model. 
Hyper-parameters may include width, depth, shortcuts, convolutional layers, activation functions, etc. 
In autoML, some systems allow neural network models to grow by adding more convolutional blocks.
In progressive systems, neural network models grow by adding neurons and links. 
Alternatively, neural network models can evolve by adding or deleting sentences and commands in programming scripts. 
These system designs are specified by users and user-specific influence is represented by $\theta_1$ in the formula. 

%theta2
%hyper-parameters that determines number of parameters in the model, functions of parameters, parameters of interest, trivial parameters
$\theta_2$ represents the hyper-parameters in neural network models, which set the number of parameters, key parameters, trivial parameters, etc. 
Hyper-parameters shape the structure of neural network models, and they are optimized in autoML by searching for the network with the best performance. 
Given hyper-parameters, parameters are optimized through the back-propagation of loss gradients. 
%theta3
%parameters, parameters of interest, trivial parameters
$\theta_3$ represents parameters in neural network models including both parameters of interest and trivial parameters. 
%theta1 determines elements in theta2
We can see that $\theta_1$ sets elements in $\theta_2$, which determines elements in $\theta_3$.
%theta2 determines elements in theta3
In this way, we divide the unknown parameters in autoML into three groups: tacit knowledge level $\theta_1$, hyper-parameters level $\theta_2$, parameters level $\theta_3$. 
$\theta_1$ and $\theta_2$ together have an influence upon the generalization capability of neural network models. 
To consider vastly different tasks, greater changes should be made to neural network models and varying $\theta_1$ and $\theta_2$ will allow this to happen.   
With the optimization of user-specific tacit knowledge $\theta_1$ considered, general AI algorithms show better generalization capability and the best performance achievable by each user is improved. 

%in model search of autoML, theta2 (model architecture) is optimized, automated search for optimal model structure, with evolutionary algorithm from zero model, find model architecture with best performance
In the process of model search in autoML, $\theta_2$ (model structure) is optimized through automated searches for the optimal model with the best performance. 
Searching for the optimal hyper-parameter value is time-consuming. 
At each value of hyper-parameters, a neural network model is trained to estimate $\theta_3$ and to obtain its performance measure. 
Here we use autoML to describe these three components of parameters in detail.
In practice, unknown parameters in any general AI algorithm can be written as
\[
\theta=(\theta_1,\theta_2,\theta_3),
\]
$\theta_1$: User-specific tacit knowledge.

$\theta_2$: Hyper-parameter for model structure.

$\theta_3$: Parameters.

where $\theta_1$ is the user characteristics representing tacit knowledge from user-specific comprehension of task information and technical preferences of users, $\theta_2$ is the hyper-parameter of general AI architecture set by $\theta_1$, $\theta_3$ is the parameter in proposed AI algorithms for tasks. 

%in model training under each hyper-parameter combination, theta3 is optimized, find parameters in best model architecture
%tacit knowledge theta1 cannot be determined automatically in algorithms
Tacit knowledge is represented by user characteristics $\theta_1$ that determine user interpretation of task information and habits of model design.
Under causal inference framework, $\theta_1$ is optimized by identifying the optimal user characteristics that correspond to the best performance of models designed by the user. 
%tacit knowledge is based upon user understanding of task information, related to user characteristics, can be dealt with causal inference framework
Traditionally tacit knowledge is not accounted for in general AI algorithms. 
In this manuscript, causal inference is applied to consider optimization with respect to tacit knowledge. 
%user-oriented, tacit knowledge, these hyper-parameters, or hyper-hyper-parameters
Through application of user-oriented smart general AI systems, models constructed by each user are improved on all tasks.

\subsection{Optimization of User-Oriented Hyper-Parameters}
\label{Optimization-of-User-Oriented-Hyper-Parameters}

\begin{figure}[htpb]
	\centering
	\includegraphics[width=0.4\linewidth]{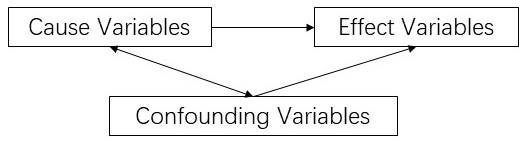}
	\caption{Cause Variables. Effect Variables. Confounding Variables.}
	\label{p2}
\end{figure}

Figure \ref{p2} illustrates the relation between cause variables, effect variables and confounding variables. 
Causal relations can be represented using directed acyclic graphs \cite{Robins2003} of node variables. 
Cause variables lead to changes in effect variables of interest. 
Confounding variables are difficult to measure since they affect both cause and effect variables, thus have an impact on the magnitude of causal effect.
It is assumed that all confounding variables are observed in data. 
Otherwise magnitude estimation of causal effect is subject to harmful bias induced by unknown unobserved factors. 
After specifying the causal model architecture with cause variables, effect variables and confounding variables defined, 
the optimal values of cause variables can be estimated which lead to the best values of effect variables conditional upon confounding variables. 

It is difficult to identify all confounding variables within tasks, especially in real scenarios with complex varying environmental conditions. 
Under causal inference framework, SUTVA (Stable Unit Treatment Value Assumptions) \cite{Rubin1986} assumptions are central to the optimization of user-oriented hyper-hyper-parameters $\theta_1$. 
%double checked with other papers with sutva words online, correct citation
%sutva
%stable unit treatment value assumptions
In SUTVA, first, it is assumed that the magnitude of causal effect is stable over individuals. 
Second, consistency assumption implies that there is no unobserved confounding variable in causal relations. 
Third, cause variable values of one unit are not influenced by other units. 
Under SUTVA assumptions, magnitude of causal effect can be estimated consistently with observed results from data. 
We can search through all values of user characteristics to identify the optimal value corresponding to the best performance of models designed by the user. 
Next section summarizes causal inference methods to estimate the optimal user characteristics from data in memory modules concerning model training experiences of many users. 

\subsubsection{Data and Framework}
\label{Data-and-Framework}

There are two primary frameworks in causal inference to compute the optimal value of user characteristics: Q learning \cite{Watkins1992} and A learning \cite{Schulte2014}.
Denote $\theta_1$ as the user characteristics that can be manipulated to improve acquired tacit knowledge and thus improve model performance. 
In observed data from memory modules, user characteristics are written as
$\bar{\theta}_{1k}=(\theta_{11},\theta_{12},\cdots,\theta_{1k})$, which consist of $k$ user characteristics for $n$ individuals. 
For example, for the $i$th user, observed user characteristics are denoted as
$\bar{\theta}_{1ki}=(\theta_{11i},\theta_{12i},\cdots,\theta_{1ki})$.
These user characteristics describe features of tacit knowledge exploited by users at different stages of data modeling and are treated as cause variables.
Effect variable is the performance of designed models.
Other individual features and task features related to both cause and effect variables are regarded as confounding variables. 
The ultimate potential outcome of model performance should be optimized to issue recommendations upon user characteristics:
\[
\max_{\bar{\theta}_{1K}} Y^*(\bar{\theta}_{1K}),
\]
where $Y^*(\bar{\theta}_{1K})$ is the potential outcome of model performance given user characteristics after considering all confounding variables.
In memory modules, observed data include user characteristics $\bar{\theta}_{1K}$, trained model performance $Y(\bar{\theta}_{1K})$ and confounding variables of many users. 
Observed data of different individuals are asssumed to be independently identically distributed (i.i.d.), where user heterogeneity is captured within observed individual features. 
From memory modules, a projection is estimated from user characteristics $\bar{\theta}_{1K}$ to observed model performance $Y(\bar{\theta}_{1K})$. 
Under SUTVA assumptions, $Y(\bar{\theta}_{1K})$ is an unbiased estimator of $Y^*(\bar{\theta}_{1K})$.
Then recommendations upon user characteristics are provided by
\[
\max_{\bar{\theta}_{1K}} Y(\bar{\theta}_{1K}).
\]
There are many approaches to find the global optimum such as random search, grid search, genetic algorithm based search, reinforcement learning based search, Bayesian sampling based search, etc. 
Tacit knowledge applied in general AI is subject to human influence and can be represented with user characteristics.
Since user characteristics are not high-dimensional, 
optimization with respect to user characteristics can be conducted efficiently.

From another perspective, user characteristics are ordered along the crucial decision points in data modeling and can be optimized sequentially. 
User characteristics at later stages depend upon user characteristics at ealier stages so that intermediate performance evaluations rely upon all prior user characteristics.
The corresponding potential outcomes of $K$ user characteristics are
\[
W^*=\{S^*_2(\theta_{11}),S^*_3(\bar{\theta}_{12}),\cdots,S_k^*(\bar{\theta}_{1k-1}),S_{k+1}^*(\bar{\theta}_{1k}),\cdots,S^*_K(\bar{\theta}_{1K-1}),Y^*(\bar{\theta}_{1K})\},
\]
where $S_{k+1}^*(\bar{\theta}_{1k})$ are the additional extra intermediate model performance measure based upon the $k$th user characteristic.
Our goal is to optimize the ultimate model performance $Y^*(\bar{\theta}_{1K})$ and identify the corresponding user characteristics $\bar{\theta}_{1K}$ as individualized recommendations. 
In memory modules, observed data include user characteristics $\theta$, intermediate performance measures $S$, and trained model performance $Y$: $(S_1,\theta_{11},S_2,\theta_{12},\cdots,S_K,\theta_{1K},Y)$.
For different users, observed data in memory modules are i.i.d. and satisfy SUTVA assumptions. 
For example, for the $i$th user, observed data is denoted as
$(S_{1i},\theta_{11i},S_{2i},\theta_{12i},\cdots,S_{Ki},\theta_{1Ki},Y_i)$.
From memory modules, projections from user characteristics $\bar{\theta}_{1k}$ to intermediate model performance measures can be estimated. 
Under SUTVA assumptions, estimated projections are consistent estimators of potential outcomes for performance measures. 

Potential model performance measures can be estimated using data from memory modules including user characteristics, confounding variables and model performances of many users. 
Maximizing estimators of potential model performance produces recommendations on the optimal value of user characteristics.
Denote $\bar{s}_k=(s_1,s_2,\cdots,s_k)$, where $s_k$ is an unbiased estimator of $S_k^*(\bar{\theta}_{1k-1})$ using observed data from memory modules.
The optimal value of user characteristics is computed sequentially from the last stage to the first stage. 
At the final stage, the optimal user characteristic is from the maximization of potential model performance: 
\[
\theta^*_{1K}(\bar{s}_K,\bar{\theta}_{1K-1})
=\text{argmax}_{\theta_{1K}} \mathbb{E}[Y^*(\bar{\theta}_{1K-1},\theta_{1K})|\bar{s}_K],
\]
and the best achievable ultimate model performance is
\[
V^*_{K}(\bar{s}_K,\bar{\theta}_{1K-1})
=\max_{\theta_{1K}} \mathbb{E}[Y^*(\bar{\theta}_{1K-1},\theta_{1K})|\bar{s}_K].
\]
At earlier stages, the optimal user characteristics are from the maximization of the expected intermediate performance measure at the next stage:
\[
\theta^*_{1k}(\bar{s}_k,\bar{\theta}_{1k-1})
=\text{argmax}_{\theta_{1k}} \mathbb{E}[V^*_{k+1}(\bar{s}_k,S^*_{k+1},\bar{\theta}_{1k-1},\theta_{1k})|\bar{s}_k],
\]
and the intermediate performance measure is optimized to be
\[
V^*_{k}(\bar{s}_k,\bar{\theta}_{1k-1})
=\max_{\theta_{1k}} \mathbb{E}[V^*_{k+1}(\bar{s}_k,S^*_{k+1},\bar{\theta}_{1k-1},\theta_{1k})|\bar{s}_k].
\]
At the first stage, the optimal user characteristic is
\[
\theta^*_{11}(s_1)
=\text{argmax}_{\theta_{11}} \mathbb{E}[V^*_2(s_1,S^*_{2},\theta_{11})|s_1],
\]
and the first intermediate performance measure is optimized to be
\[
V^*_{1}(s_1)
=\max_{\theta_{11}} \mathbb{E}[V^*_2(s_1,S^*_{2},\theta_{11})|s_1].
\]
We can see that estimators of potential outcomes must be unbiased in order for this recursive process to proceed smoothly. 
This is why we need SUTVA assumptions to hold.  
Otherwise, bias accumulates along sequential optimization and eventually destroys the hope of identifying the correct optimal value of user characteristics. 
By optimization from the last stage to the first, estimated optimum is closer to the global optimum. 

\subsubsection{Q Learning}
\label{Q-Learning}
%\cite{Watkins1992}
%q learning

Q learning is proposed in \cite{Watkins1992} and summarized in \cite{Schulte2014} as an important method to estimate the optimal value of cause variables in the context of causal inference. 
Q learning is concentrated upon estimating Q functions, which are defined to be projections from user characteristics to model performance measures after considering confounding variables. 
Then Q functions are inserted into causal inference framework to perform sequential optimization and to estimate the optimal user characteristics as recommendations from user-oriented general AI systems. 
At the final stage, Q function is defined as the expected model performance:
\[
Q_K(\bar{s}_K,\bar{\theta}_{1K})=\mathbb{E}(Y|\bar{s}_K,\bar{\theta}_{1K}).
\]
Under SUTVA assumptions, maximization of expected potential outcomes amounts to the maximization of Q functions. 
It is critical that Q functions are unbiased estimators of the expected model performance measures.
Otherwise, bias accumulates in sequential optimization and the recommended user characteristics are not helpful for the improvement of model performance. 
The optimal user characteristic at the final stage is from the maximization of Q function at the last stage:
\[
\theta^Q_{1K}=\text{argmax}_{\theta_{1K}} Q_K(\bar{s}_K,\bar{\theta}_{1K-1},\theta_{1K}).
\]
The optimal ultimate model performance at the final stage is
\[
V^Q_K=\max_{\theta_{1K}} Q_K(\bar{s}_K,\bar{\theta}_{1K-1},\theta_{1K}).
\]
At earlier stages, Q functions are defined as the expected model performance at the next stage:
\[
Q_k(\bar{s}_k,\bar{\theta}_{1k})=\mathbb{E}[V^Q_{k+1}(\bar{s}_k,s_{k+1},\bar{\theta}_{1k})|\bar{s}_k,\bar{\theta}_{1k}].
\]
The optimal user characteristics at earlier stages are from the maximization of expected model performance at the next stage:
\[
\theta^Q_{1k}=\text{argmax}_{\theta_{1k}} Q_k(\bar{s}_k,\bar{\theta}_{1k-1},\theta_{1k}).
\]
The optimal intermediate performance measures are
\[
V^Q_k=\max_{\theta_{1k}} 
Q_k(\bar{s}_k,\bar{\theta}_{1k-1},\theta_{1k}).
\]
In Q learning, to estimate Q functions, regression methods are often applied. 
Neural networks and other machine learning models can also be utilized to estimate the mapping from user characteristics to performance of designed models.
Q functions are posited to be $Q_k(\bar{s}_k,\bar{\theta}_{1k};\eta_k)$, where $\eta_k$ is the unknown parameter to be estimated. 
For example, in neural network based Q functions, $\eta_k$ comprises weights and bias parameters, which are estimated through back-propagation of loss gradients. 
In regression based Q functions, $\eta_k$ consists of coefficients and variance parameters, which can be computed using estimating equations. 
For users $i=1,2,\cdots,n$, estimating equations of $\eta_k$ can be written as
\[
\sum_{i=1}^n \frac{\partial Q_k(\bar{s}_k,\bar{\theta}_{1k};\eta_k)}{\partial \eta_k}
\Sigma_k^{-1}(\bar{s}_k,\bar{\theta}_{1k})\times
[V^Q_{k+1i}-Q_k(\bar{s}_k,\bar{\theta}_{1k};\eta_k)]=0,
\]
where the solution to this equation is the estimator of $\eta_k$. 
There is no specification of Q function models that outperforms all others. 
The performance of Q learning substantially relies upon the quality of Q function models. 
After computing the optimal value of user characteristics, we should use variability of estimated parameter $\hat{\eta}_k$ to evaluate the reliability of estimated results.
Inference upon $\eta_k$ includes evaluations of bias, noise and variability in $\hat{\eta}_k$, which affect precision and consistency of the estimated optimal value of user characteristics.
Furthermore, correctness of posited Q functions models should be examined and robustness of estimators should be analyzed under possible misspecifications of Q function models \cite{Unknowng}.  

\subsubsection{A Learning}
\label{A-Learning}

%\cite{Schulte2014}
%a learning
Another mainstream method to estimate the optimal value of cause variables is A learning \cite{Schulte2014}. 
%optimization from ending to begining
%from last hope to initial hope
%estimation of value function
%specified model for value and advantage functions
%
Q function models are posited to estimate projections from user characteristics to performance measures after considering confounding variables. 
Q functions represent the expected outcomes at different values of cause variables. 
To identify the optimal value of cause variables that corresponds to the best outcome, Q functions are maximized.
From another perspective, we can compute differences in expected outcomes at different values of cause variables to identify the better value. 
In A learning, models are posited upon the difference in expected outcomes rather than the values of expected outcomes. 
A learning is usually more efficient than Q learning in the estimation of optimal cause variable values.
However, when cause variables can take a large number of different values, pairwise comparison of expected outcomes is not feasible. 

For example, when cause variable is binary and takes values of either 1 or 0, regression based Q functions are posited as follows:
\[
Q_1(s_1,\theta_{11};\eta_1)=s_1\eta_{11}+\theta_{11}\eta_{12}+s_1\theta_{11}\eta_{13},
\]
where $\theta_{11}$ is the binary cause variable and $\theta_{11}=0$ or $\theta_{11}=1$, $\eta_1=(\eta_{11},\eta_{12},\eta_{13})$ is the unknown parameter to be estimated \cite{Unknowng}. We can see that
\[
Q_1(s_1,\theta_{11}=1;\eta_1)=s_1\eta_{11}+\eta_{12}+s_1\eta_{13},
\]
and that
\[
Q_1(s_1,\theta_{11}=0;\eta_1)=s_1\eta_{11},
\]
then the difference in expected outcomes at different cause variable values is 
\[
Q_1(s_1,\theta_{11}=1;\eta_1)-Q_1(s_1,\theta_{11}=0;\eta_1)=\eta_{12}+s_1\eta_{13}.
\]
When $\eta_{12}+s_1\eta_{13}>0$, we know that $Q_1(s_1,\theta_{11}=1;\eta_1)>Q_1(s_1,\theta_{11}=0;\eta_1)$ and $\theta_{11}=1$ is the optimal cause variable value. Otherwise, $\theta_{11}=0$ is the optimal cause variable value.
In Q learning, models are posited on Q functions, and $\eta_{11},\eta_{12},\eta_{13}$ should be estimated.
In A learning, models are specified on the differences in expected outcomes, and $\eta_{12},\eta_{13}$ should be estimated.
In a nutshell, A learning contains less number of unknown parameters and is generally more efficient when cause variables are binary. 

In A learning, differences in expected outcomes (Q functions) are termed as contrast functions. 
In the broad sense, contrast functions are the projections from user characteristics and cause variables to differences in expected outcomes after considering all confounding variables.
Models specified for contrast functions can be neural networks, linear or nonlinear regressions, etc. 
A simple example of estimating equation for regressional contrast function model \cite{Schulte2014} at the $k$th stage is
\[
\sum_{i=1}^n \lambda_k(\bar{s}_k,\bar{\theta}_{1k-1};\psi_k)\{
\theta_{1k}-\pi_k(\bar{s}_k,\bar{\theta}_{1k-1};\phi_k)\}\times
\{V^Q_{k+1}-\theta_{1k}C_k(\bar{s}_k,\bar{\theta}_{1k-1};\psi_k)
-h_k(\bar{s}_k,\bar{\theta}_{1k-1};\beta_k)\}=0,
\]
\[
\sum_{i=1}^n \frac{\partial h_k(\bar{s}_k,\bar{\theta}_{1k-1};\beta_k)}{\partial \beta_k}
\times \{V^Q_{k+1}-\theta_{1k}C_k(\bar{s}_k,\bar{\theta}_{1k-1};\psi_k)
-h_k(\bar{s}_k,\bar{\theta}_{1k-1};\beta_k)\}=0,
\]
where $\lambda_k(\bar{s}_k,\bar{\theta}_{1k-1};\psi_k)=\partial C_k(\bar{s}_k,\bar{\theta}_{1k-1};\psi_k)/ \partial \psi_k$, $C_k(\bar{s}_k,\bar{\theta}_{1k-1};\psi_k)$ is the model posited for contrast function,
$\pi_k(\bar{s}_k,\bar{\theta}_{1k-1};\phi_k)=P(\theta_{1k}=1|\bar{s}_k,\bar{\theta}_{1k-1})$ is computed empirically from observed data,
$h_k(\bar{s}_k,\bar{\theta}_{1k-1};\beta_k)$ is the posited baseline model.
Under the misspecification of either contrast model or baseline model, estimated optimal user characteristic $\theta_{1k}$ is still valid.  
For brevity, user characteristic $\theta_{1k}$ is considered to be binary and to choose values from only two scalars. 
In UOGASuCI, user characteristics may take more than two distinct values. 
Contrast function models based upon regressions in these estimating equations are not practically feasible when user characteristics are not binary.

%select user characteristics that are associated with best model performance found by that individual
Since numerous user characteristics are collected or extracted from model training experiences of many users, 
model selection can be applied to identify the set of user characteristics that are significantly associated with the best performance of models designed by the user \cite{Unknowng}. 
%since lots of user characteristics are collected
In addition, it is necessary to find all confounding variables of cause and effect variables, as in figure \ref{p2}. 
When parameter inference is conducted after model selection, randomness in the selected model should be taken into account in deriving reliability measures of estimated results, as in post-selection inference. 
In neural network models, feature extraction modules perform dimension reduction and model selection, other layers further process data and produce output. 
The whole neural network is trained jointly using back-propagation of loss gradients. 
The process of model selection is data-driven and is influenced by the randomness in original data. 
As a consequence, parameter inference in neural network should also consider applying post-selection inference. 

As in figure \ref{p1}, in external memory modules, we save model training experiences of many users and diverse tasks. 
In the task dimension, we utilize meta-learning to identify the most analogous task model as the proposed initial model for an unseen task. 
Then fine-tuning guided by meta-learner is applied to update base learners more efficiently. 
In the user dimension, we apply causal inference to devise UOGAS, where tacit knowledge applied in general AI system is represented by a set of user characteristics. 
In addition, other user-specific characteristics that may affect performance of models designed by this user are also included. 
This set of user characteristics pertaining to user model design is treated as the cause variable. 
The best performance of models designed by each user is regarded as the effect variable. 
We also record all confounding variables that may affect both cause and effect variables. 
Under the context of causal inference, we identify the optimal value of user characteristics that lead to the best performance of devised models by the user. 
In the user dimension, we recommend the optimal value of user characteristics and it is up to the user to decide whether to take these suggestions or not. 

Q learning and A learning are notable frameworks in causal inference to estimate the optimal value of cause variables. 
They can be integrated into UOGAS to perform individualized optimization in order to boost the performance of models designed by each user. 
There are also other platforms to derive user-oriented optimization of user characteristics to achieve the best individualized model design. 
Under the framework of causal inference, user-oriented general AI system is termed as UOGASuCI, where uCI means 'under causal inference'. 
With other platforms embedded, user-oriented systems can be termed accordingly. 
	
\section{Neural Network with Causal Inference}	
\label{Neural-Network-with-Causal-Inference}

%network with causal links
Links in neural networks can be specified as fixed or random \cite{Lochmann2011} by model design. 
%in convolutions, links are averages, linear combination of components in average, linear associations
In convolutional layers, links are fixed and constitute moving averages, which are linear combinations of original data with unknown weight parameters to be estimated. 
%in temporal convolutions, links are past influence upon future, present depends only upon history information
In temporal convolutions, links represent past influence upon future, and are moving averages of previous data points. 
%links can be randomly trimmed to reduce over-fit
In dropout technique, links follow a probability distribution, where links are trimmed at random to alleviate model overfit. 
%random links, infuse randomness into links, for each node, represent model selection, condition upon selected model, nodes are correlated, selected model for each node is similar, no need to conduct model selection at each node, since node number is large in deep network model, input dimension is high, 
When randomness is infused into links, information from previous layer is censored by random selection. 
Since nodes in neural networks are often highly correlated, weight parameters are often shared by different nodes located close to each other. 
As input dimension grows, the number of nodes and weight parameters also increase in neural network models. 
There is no need to conduct random selection on links at every node, and random selection at one node can be directly replicated at other nodes. 

%over-emphasize on the part of interest
In the design of model architecture, we over-emphasize on the region of interest and bias our model design towards that part in the original data. 
We focus our attention upon the main objective and ignore components that we think are trivial in model design.
%overlook parts not of interest, but contains information
We judge whether each component is useful or not based upon our model training experiences using our tacit knowledge about task information. 
%links represent associations between current layer nodewise information and previous layer nodewise information
Links in neural network represent associations between nodes in current layer and nodes in previous layers. 
Links are concentrated upon the region of interest and are designed to extract more information from target areas. 
Links represent model selection upon variants and information censoring upon processed data. 
Link designs are influenced by tacit knowledge and user characteristics, and can be optimized using causal inference framework. 
Random selection in links is purely random but the probability distribution of links are specified by users and influenced by user characteristics.
%
%activation function in neural network
%randomness in activation
Similarly, randomness can be infused into activation functions in neural networks. 
But back-propagation of loss gradients through random links and random activation functions is more complex than fixed cases. 
%randomness in linear combination of last layer node information, randomness in projection

%like in statistical inference, model is trained, parameter inference (hypothesis testing and confidence) is provided after model is trained, based upon estimated parameters and corresponding variance estimate
In statistical inference, a linear model is trained. 
Afterwards parameter inference (hypothesis testing and confidence interval) is provided based upon estimated parameters and the corresponding variance estimate.
%deterministic information flow, deterministic gradient propagation, efficient for objective-based optimization of neural network model
%undeterministic information flow, gradient propagation impossible, not good for training models from optimization, can be used in pre-trained deep model, to modify model from perspective of inference
By analogy with statistical inference, a deep neural network is pre-trained and causal inference is performed upon weight parameters in the pre-trained model. 
Under the framework of Q learning and A learning, the general AI system can decide whether links are significant contributions to better performance or not.
If not, these links can be removed.  
%in pre-trained model, inference on each weight, and on each activation function, whether null or not, assume undeterministic information flow, these quantities need inference, re-build network model after training using inference results
%likewise, in linear model, 
%y=xb+e
%parameter estimate
%hypothesis test
%confidence interval
In some cases, removing links leads to better model performance. 
In other cases, keeping links there brings better results. 
%we use causal inference to conduct such inference jobs
Based upon inference, we can re-build the network from modifications of the pre-trained model. 

Statistical inference using pre-trained deep model is intractable. 
%in neural network model,
For example, a neural network model is written as
%y=f(x,theta)+e
\[
y=f(x,\theta)+\epsilon,
\]
where $y$ is the data annotation, $x$ is the input data, $\theta$ is the unknown weight and bias parameter, $\epsilon$ is the random error.
%loss gradient backpropagation
%estimate theta
$\theta$ is estimated using back-propagation of loss gradients. 
%since deep model has very high prediction accuracy
%variance of e is small
Since the pre-trained deep model has very high prediction accuracy, variance of $\epsilon$ is a small positive quantity and very close to $0$.
%hypothesis test and confidence interval of theta not available
Therefore hypothesis test and confidence interval of $\theta$ are not available. 
%confidence interval is very tight, point estimate of theta
Confidence interval of $\theta$ is very tight and close to a point estimate of $\theta$.
We cannot test whether a link is significant or not based upon pre-trained deep model. 
%no room for statistical inference
%for each link, infer whether link should be there
But under causal inference context, we can compare model performances with the link and without the link.
We make the optimal choice leading to the best model performance. 
%does remove this link lead to better performance
%

\section{Conclusion and Discussion}
\label{Conclusion-and-Discussion}

%tacit knowledge
Traditionally, it is thought that tacit knowledge is user-specific and cannot be optimized automatically in algorithms. 
%individualized comprehension of task information
Tacit knowledge has a significant influence upon model performance but cannot be described explicitly as a component in algorithms. 
Tacit knowledge refers to individualized comprehension of task information, user-specific modeling preferences, personal bias in designing model architecture, etc. 
%hard to account for in algorithms
%this paper proposes causal inference framework to consider tacit knowledge in general AI
This paper proposes UOGAS, in which causal inference framework is applied to account for tacit knowledge in general AI. 
The set of user characteristics that determine tacit knowledge is found. 
The optimal value of these user characteristics is identified through causal inference and suggested to users. 
At optimum, performance of models designed by the user is expected to be better than before. 
For example, recommendations for me will be doing more programming and working more on real data. Recommendations for high-school students will be obtaining more computer science courses. Suggestions for experienced engineers will be more professional such as increasing exposure to unsupervised learning. 
The optimal value of user characteristics is a personalized recommendation to help users design better models. 
%other ways to consider tacit knowledge
%information, logical reasoning, subjective and objective judgments from data
There are also other ways to model tacit knowledge and perform user-oriented optimization, such as logical reasoning, knowledge hierarchy to represent subjective and objective judgments from data.
%knowledge hierarchy
UOGAS is an example to consider human-AI interaction, where AI assists human in designing better AI. 

A thorough research framework is established in \cite{Roberts2021} pointing out principles concerning deep neural network theories. Tacit knowledge represented by user characteristics may be quantified with mastering level of these principles in \cite{Roberts2021}.
How to measure these user characteristics and what dimensions to focus on remain open problems for future research. 
Ideally these user characteristics are exactly the set of all features related to user-specific modelling issues, which constructs a user-specific mapping from data to its optimized algorithmic model. 

Humans are better at high-quality creative work, and machines are better at repeated jobs, which can be improved dramatically through sufficient training. Can innovation capability be acquired through training? We propose an open framework to combine humans and machines into one system, where they help each other to build better models. In practice, 
people rely upon both humans and machines results to make the final decision. As a consequence, these human-machine joint models are considered to optimize the process of human-machine interaction for better algorithms. 

GLOM \cite{Hinton2021} allows deep neural network models to partition high-dimensional images into image-specific part-whole hierarchies. 
Perceiver \cite{Jaegle2021} builds upon deep transformer models to simultaneously process high-dimensional multi-modal data. Perceiver iteratively analyzes multi-modal data to identify the most related information. 
Both GLOM and Perceiver introduce higher level of flexibility into deep neural network structure, so that network is more adaptive to diverse tasks from multiple modalities.

\section*{Acknowledgment}

I appreciate valuable comments from Jacob Valdez (jacob.valdez@limboid.ai).

\bibliographystyle{unsrt}

%\bibliography{appl2}


\end{document}